\documentclass{article}
\usepackage{spconf,amsmath,epsfig}
\usepackage{graphicx}
\usepackage{subcaption}
\usepackage{indentfirst}
\usepackage{threeparttable}

\let\OLDthebibliography\thebibliography
\renewcommand\thebibliography[1]{
  \OLDthebibliography{#1}
  \setlength{\parskip}{0pt}
  \setlength{\itemsep}{0pt plus 0.3ex}
}

\begin{document}\sloppy

\def\x{{\mathbf x}}
\def\L{{\cal L}}

\title{SemiPL: A Semi-supervised Method for Event Sound Source Localization}
\name{Yue Li$^1$, Baiqiao Yin$^1$, Jinfu Liu$^1$, Jiajun Wen$^1$, Jiaying Lin$^1$, Mengyuan Liu$^\dagger$$^2$}
\address{$^1$School of Intelligent Systems Engineering, Sun Yat-sen University \\
$^2$National Key Laboratory of General Artificial Intelligence, Peking University, Shenzhen Graduate School\\
Corresponding author is Mengyuan Liu (e-mail: liumengyuan@pku.edu.cn). }

\maketitle

\maketitle

\begin{abstract}
In recent years, Event Sound Source Localization has been widely applied in various fields. Recent works typically relying on the contrastive learning framework show impressive performance. However, all work is based on large relatively simple datasets. It's also crucial to understand and analyze human behaviors (actions and interactions of people), voices and sounds in chaotic events in many applications, e.g., crowd management, and emergency response services. In this paper, we apply the existing model to a more complex dataset, explore the influence of parameters on the model, and propose a semi-supervised improvement method SemiPL. With the increase in data quantity and the influence of label quality, self-supervised learning will be an unstoppable trend. The experiment shows that the parameter adjustment will positively affect the existing model. In particular, SSPL achieved an improvement of 12.2\% cIoU and 0.56\% AUC in Chaotic World compared to the results provided. The code is available at:
https://github.com/ly245422/SSPL
\end{abstract}

\section{Introduction}
Vision can influence our perception of sound. For example, when we view a picture, visual images can change our interpretation and emotional experience of the sound. Visual images, colors, and motion can guide our perception of sound and the construction of scenarios. Hearing can also influence our perception and experience of visuals \cite{gaver1993world}. For example, when we hear a sound, the pitch, volume, and rhythm of the sound can influence our perception and emotion of visual objects. Sound can enhance or diminish our attention, emotions, and feelings about our visual environment. 

When we hear the roar of a car engine, we can infer the speed and motion of the car based on the pitch and volume of the engine sound. Early experiments \cite{frassinetti2002enhancement} have shown that sudden sounds enhance perceptual processing of subsequent visual stimuli. This sound perception of speed directly affects our visual perception of the road and vehicles around us. This interplay and interaction stems from the brain's ability to synthesize and process different sensations. By integrating existing sensory inputs, the brain develops a more comprehensive and accurate perceptual experience. This interaction also reflects the connections and coordination between perceptions, helping us to better understand and adapt to our surroundings.

Thus, acoustic event detection is a rather broad topic \cite{hershey1999audio} in the field of environmental sound detection and classification, with a wide range of applicability in surveillance and monitoring, assistive technologies, and multimedia indexing. Recent works showing impressive localization performance typically rely on the contrastive learning framework. With the increase in data quantity and the influence of label quality, self-supervised learning will be an unstoppable trend in the future \cite{van2020survey}. For datasets with partial labels, undoubtedly, semi-supervised learning is the best choice and also the inevitable trend for the future development of sound source localization. So, we propose a semi-supervised method SemiPL.

The main contributions of this paper are:
\begin{itemize}
    \item We apply SSPL to the Chaotic World dataset achieving an improvement of 12.2\% cIoU and 0.56\% AUC.
    \item We explore the application of the semi-supervised method SemiPL and the effect of different parameters on SSPL on the Chaotic World dataset.
\end{itemize}

\section{Related Work}
\subsection{Sound Localization in Visual Scenes}
Sound source localization within visual scenes seeks to pinpoint the location of objects producing sound within a specified image. Initial efforts to address this challenging task predominantly focused on leveraging statistical modeling of crossmodal relationships, utilizing techniques such as mutual information and canonical correlation analysis to effectively capture the underlying associations. Nonetheless, these shallow models primarily demonstrate their strengths in relatively straightforward audio-visual scenarios. Currently, employing deep learning techniques to tackle this task has become the mainstream approach. For example, Senocak et al. \cite{senocak2018learning} introduce an innovative unsupervised algorithm for localizing sound sources by utilizing an attention mechanism. This mechanism is guided by auditory information in conjunction with paired sound and video frames, offering a sophisticated approach to address the problem. In response to the daunting challenge of visually localizing multiple sound sources in unconstrained videos without the availability of pairwise sound-object annotations, Qian et al. \cite{qian2020multiple} engineered a two-stage audiovisual learning framework. This innovative solution separates audio and visual representations of different categories within complex scenes and subsequently executes a cross-modal feature alignment in a progressively refined manner.
\subsection{Self-Supervised Visual Representation Learning}
Self-supervised learning (SSL) has made significant strides and achieved remarkable breakthroughs in large-scale computer vision benchmarks. The majority of contemporary SSL methods rely on the implementation of contrastive learning strategies. Fundamentally, these approaches transform a single image into multiple views, simultaneously repelling distinct images (negatives) while attracting various perspectives of the same image (positives). Numerous efforts \cite{zbontar2021barlow, ermolov2021whitening, grill2020bootstrap} have been dedicated to alleviating the reliance on negatives and streamlining the SSL framework beyond the scope of traditional contrastive learning approaches. For instance, to minimize the computational burden associated with positive and negative sample pairs, Ermolov et al. \cite{ermolov2021whitening} chart a different course by proposing a novel loss function for SSL, which is founded on the whitening of latent-space features. Grill et al. \cite{grill2020bootstrap} presents a fresh approach to self-supervised image representation learning, known as Bootstrap Your Own Latent (BYOL). This innovative method hinges on the synergy between two neural networks, dubbed online and target networks, which collaborate and continuously learn from one another.
\subsection{Audio-Visual Representation Learning.}
Often, vision and sound function as complementary senses that naturally aid in supervising audio-visual learning 
\cite{Aytar_Vondrick_Torralba_2016,Korbar_Tran_Torresani_2018,Owens_Wu_McDermott_Freeman_Torralba_2016}. For instance, in \cite{Aytar_Vondrick_Torralba_2016}, visual features drawn from a pre-trained teacher network assist a student network in acquiring more distinctive audio representations, and the reverse is also true \cite{Owens_Wu_McDermott_Freeman_Torralba_2016}. Korbar et al. \cite{Korbar_Tran_Torresani_2018} and Owens and Efros [41] leveraged the synchronization between audio and visual streams to create negative samples and formulate contrast loss, respectively, aiming to derive generalized multisensory characteristics. Certain investigations have also delved into audio-visual correspondences via feature clustering. Primarily, these methodologies concentrate on acquiring task-independent representations for downstream classification-related endeavors, including action/scene recognition, audio event categorization, and video retrieval. Nevertheless, as they are not tailored for sound source localization, their performance on this particular task remains constrained.

\section{Chaotic World Dataset}
The SSPL \cite{SSPL} model has achieved excellent results on the traditional Flickr-SoundNet \cite{aytar2016soundnet} dataset and VGG-Sound \cite{chen2020vggsound} with the leading edge. However, It's also crucial to understand and analyze human behaviors (actions \cite{liu2023novel, TDGCN, EPPNet} and interactions of people \cite{wen2023interactive}), voices, and sounds in chaotic events in many applications, e.g., crowd management, and emergency response services. Unlike human behavior in everyday life, human behavior during chaotic events is often more complex and their actions and impacts on others are often unusual. Thus, this paper aims to utilize the Chaotic World dataset \cite{ChaoticWorld}, which is a large and challenging multi-modal video dataset. 

Chaotic World dataset \cite{ChaoticWorld} consists of a total of 378,093 annotated instances for triangulating the source of sound for Event Sound Source Localization. The dataset aims to analyze multiple dimensions of a scene to study human behavior in chaotic situations in a comprehensive and detailed manner, and to understand the nature of human behavior (including human actions and interactions) during chaotic events. Sound is critical in the understanding and response to chaotic events in many applications, such as crowd management and emergency response services. Therefore, it is important to provide a comprehensive analysis of human behavior during chaotic events. 

In this paper, we use the Chaotic World dataset \cite{ChaoticWorld}, which contains data on the localization of event sound sources as well as acoustic aspects, to fully understand human behavior in chaotic situations.

\begin{figure}[h]
\centering
\includegraphics[width=0.45\textwidth]{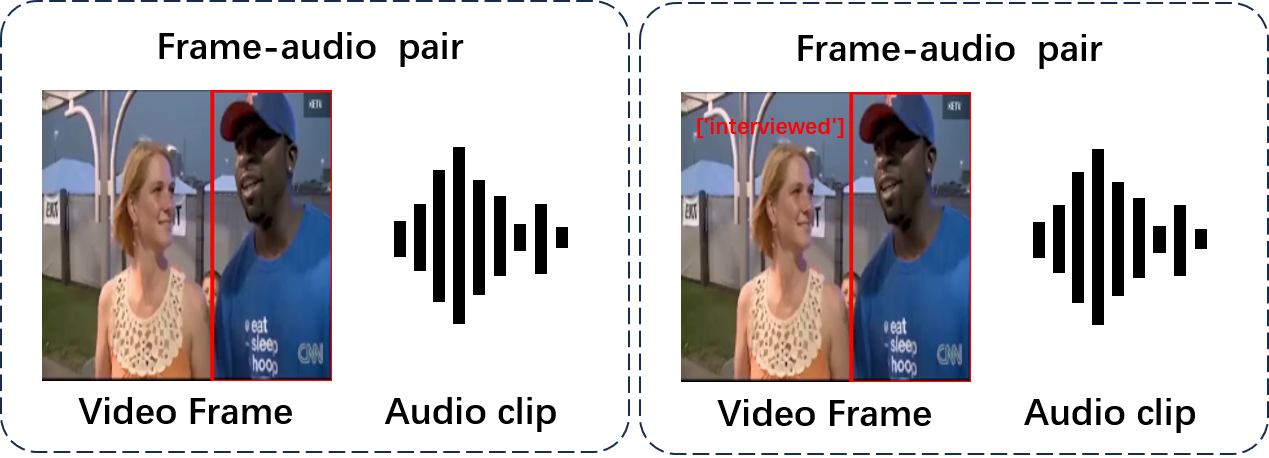}
\caption{Left: SSPL (w/ PCM) input data format. Right: SSPL (w/o PCM) input data format.}
\label{fig:dataset}
\end{figure}
\begin{figure*}[t]
\centering
\includegraphics[width=0.9\linewidth]{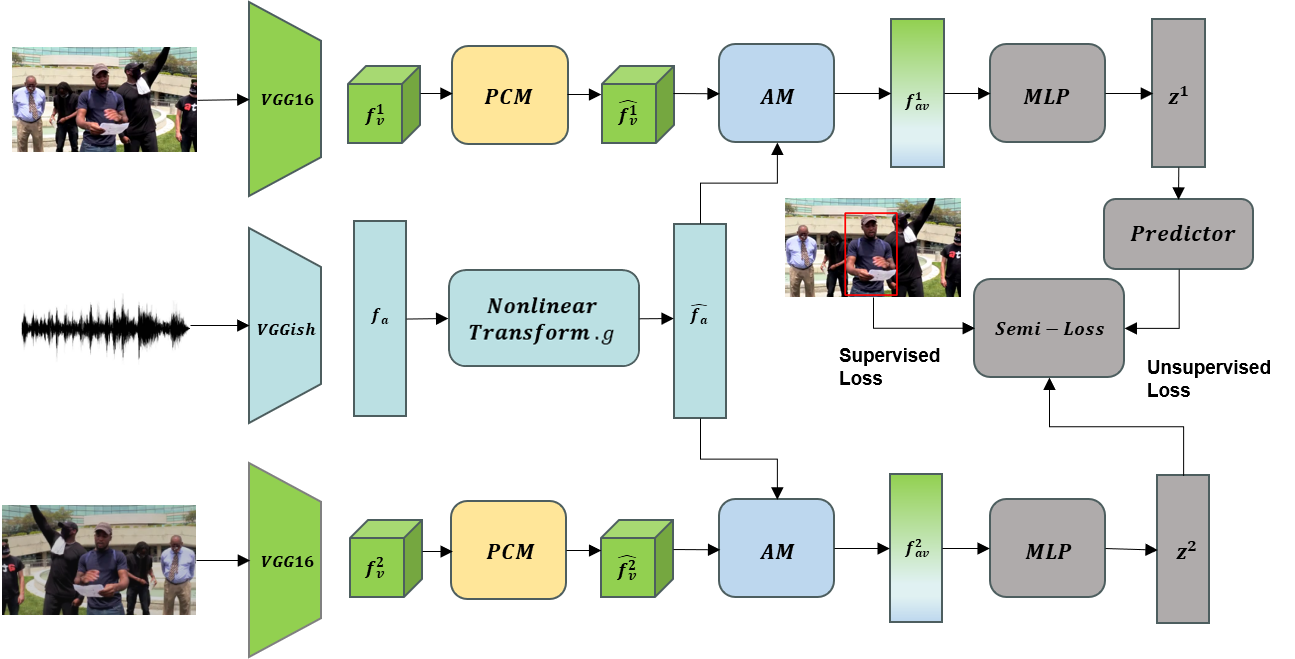}
\caption{\textbf{Framework of our Semi-Supervised Model SemiPL.} The AM and PCM modules remain consistent with SSPL, with the addition of an unsupervised loss. }
\label{fig:Semi-Supervised}
\end{figure*}
This paper uses 456 videos from this dataset, 384 training videos, and 72 test videos. We first extract each video frame by frame, and then extract the middle frame of each video and 1.5s of audio before and after the middle frame, for a total of 3s of audio, to form image-audio pairs. After that, we train the model using two random subsets of 1595 image-audio pairs and provide 304 annotated image-audio pairs for evaluation. The annotations in the Chaotic World dataset provide the fps corresponding to the start and end times of the videos and the bounding box coordinates in non-normalized xyxy format from start time to end time, we extract the intermediate video frames and the corresponding audio and annotation boxes based on the video frame index, as shown in Figure \ref{fig:dataset}. For SSPL \cite{SSPL} without PCM(a predictive coding module), we also extract the keywords ssl(type of sound) in the annotations as the class labels, as shown in Figure \ref{fig:dataset}. If there exists only one annotation frame for a certain video, the first annotation frame is selected as the true value.

\section{Method}
In this section, we delved deeper into the influence of various parameters on the effectiveness of SSPL (Self-Supervised Predictive Learning: A Negative-Free Method for Sound Source Localization in Visual Scenes) \cite{SSPL}. Through rigorous experimentation, we observed how changes in these parameters affected the algorithm's accuracy, efficiency, and stability. Additionally, we introduced our novel semi-supervised learning approach, SemiPL, which incorporates advancements in both supervised and unsupervised learning techniques. By leveraging unlabeled data more effectively, SemiPL aims to enhance the overall performance and generalizability of machine learning models, especially in scenarios where labeled data is scarce.
\subsection{Parameter adjustments}
During the research process, we found that learning rates have a significant impact on the performance of SSPL \cite{SSPL}. Therefore, we attempted to adjust the learning rates and batch size per GPU to observe their effects on the performance of SSPL \cite{SSPL}. We reduced the ssl head learning rate from 5e-5 to 3e-5. Also, we reduce batch size per GPU from 128 to 64. The results and analysis will be provided in the section experiment.
\subsection{Semi-supervised Learning}

To achieve the same performance, semi-supervised learning requires only a small amount of labeled data and a certain amount of unlabeled data, while self-supervised learning usually requires a large amount of unlabeled data \cite{zhai2019s4l}. Therefore, for the Chaotic World dataset, self-supervised learning may not obtain enough information to learn effective feature representations. We devised a semi-supervised method SemiPL under the self-supervised learning setup. We incorporated supervised loss into the SSPL \cite{SSPL} network architecture. To achieve this, we formulated a semi-supervised loss,
\begin{eqnarray}
L_{SSPL} = L_{S}\left ( \hat{y}_i, y_i \right )  + L_{U}\left ( p_{1},p_{2},z_{1},z_{2} \right ) 
\end{eqnarray}
where $L_{U}$ and $L_{S}$ denote unsupervised and supervised losses respectively. The unsupervised loss $L_{U}$ function is defined as,
\begin{eqnarray}
\text{L}_{U} = \frac{1}{2} \left( \text{NCS}(p_1, z_2) + \text{NCS}(p_2, z_1) \right)
\end{eqnarray}
where $p_{1}$ and $p_{2}$ are the outputs of the two projection heads of the network, and $z_{1}$ and $z_{2}$ are the outputs of the corresponding augmented samples, and negative cosine similarity (NCS) is calculated by:
\begin{eqnarray}
\text{NCS}(p, z) = \frac{p \cdot z}{\|p\| \|z\|}
\end{eqnarray}
$L_{S}$ utilizes the cross-entropy loss function, defined by,
\begin{eqnarray}
L_{S} = -\frac{1}{N} \sum_{i=1}^{N} \left( y_i \log(\hat{y}_i) + (1 - y_i) \log(1 - \hat{y}_i) \right)
\end{eqnarray}
where N is the size of the flattening pixel of a 224×224 image. $\hat{y}_i$ and ${y}_i$ are one-dimensional vectors obtained by flattening the ground-truth heatmap and the interpolated heatmap of the predicted values resized to 224×224 size, respectively.

The rest retains the original model structure.

\section{Experiment}
In this section, we will present the experimental results of parameter tuning and  SemiPL, analyze their strengths and weaknesses, and discuss potential improvements.
\subsection{Evaluation Metric.}
We follow \cite{LSS,SSPL}and use consensus-IoU (cIoU@0.5), whereby the score of each pixel is computed based on the consensus of multiple annotations \cite{LSS}, as well as Area Under the Curve (AUC) \cite{LSS} for Event Sound Source Localization (ESSL).

Considering that the annotations provided by the dataset are based on bounding boxes of images sized 328×120, we will first convert the bounding box annotations to size 224×224. Then, we will convert the bounding box annotations into binary maps $\left \{ gt_{i} \right \}_{i=1}^{N} $, where N is the number of subjects. Infer maps $\left \{ infer_{i} \right \}_{i=1}^{N} $ are binary maps of prediction maps, we use 0.5 for the cIoU threshold in the experiments, cIoU's formula is:
\begin{eqnarray}
\text{cIoU} = \frac{ \sum_{i=1}^{N}  \text{infer}_i \cap  \text{gt}_i }{ \sum_{i=1}^{N} \left [  \text{gt}_i  + \text{infer}_i \cup  (\text{gt}_i == 0)\right ] }
\end{eqnarray}
\begin{eqnarray}
\text{infer}_i \cap  \text{gt}_i =  \left( \text{infer}_i \times \text{gt}_i \right)
\end{eqnarray}
\begin{eqnarray}
\text{infer}_i \cup  (\text{gt}_i == 0)  =   \left( \text{infer}_i \times (\text{gt}_i == 0) \right)
\end{eqnarray}
Using the common practice in object detection, we use 0.5 for the cIoU threshold in the experiments.

\subsection{Qualitative Results and Analysis.}
\begin{figure}[h]
\centering
\begin{subfigure}{0.28\linewidth}
\caption*{Annotations}
\includegraphics[width=\linewidth]{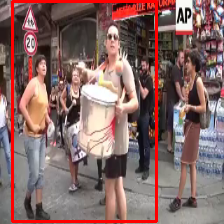}
\end{subfigure}
\hspace{0.5mm} 
\begin{subfigure}{0.28\linewidth}
\caption*{SSPL(w/ PCM)}
\includegraphics[width=\linewidth]{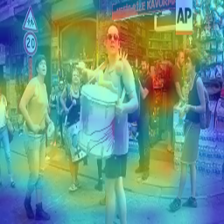}
\end{subfigure}

\begin{subfigure}{0.28\linewidth}
\includegraphics[width=\linewidth]{image/ADCCAWWD_004398_annotation.png}
\end{subfigure}
\hspace{0.5mm} 
\begin{subfigure}{0.28\linewidth}
\includegraphics[width=\linewidth]{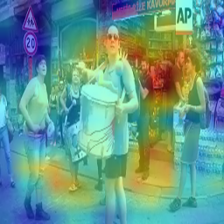}
\end{subfigure}
\caption{The first row is the self-supervised model, and the second row is the semi-supervised model SemiPL. It can be seen that self-supervised model has a somewhat larger recognition area for vocalized objects.}
\label{fig:para result}
\vspace{-5mm} 
\end{figure}

\begin{figure}[t]
\centering
\begin{subfigure}{0.28\linewidth}
\caption*{Annotations}
\includegraphics[width=\linewidth]{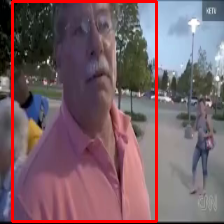}
\end{subfigure}
\hspace{0.5mm} 
\begin{subfigure}{0.28\linewidth}
\caption*{SSPL(w/o PCM)}
\includegraphics[width=\linewidth]{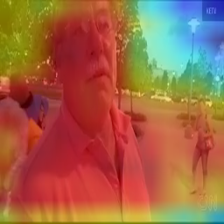}
\end{subfigure}
\hspace{0.5mm} 
\begin{subfigure}{0.28\linewidth}
\caption*{SSPL(w/ PCM)}
\includegraphics[width=\linewidth]{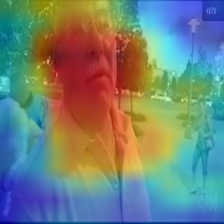}
\end{subfigure}

\begin{subfigure}{0.28\linewidth}
\includegraphics[width=\linewidth]{image/AJJFTGGI_000233_annotation.png}
\end{subfigure}
\hspace{0.5mm} 
\begin{subfigure}{0.28\linewidth}
\includegraphics[width=\linewidth]{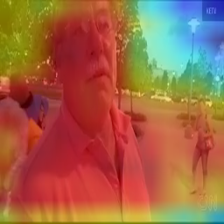}
\end{subfigure}
\hspace{0.5mm} 
\begin{subfigure}{0.28\linewidth}
\includegraphics[width=\linewidth]{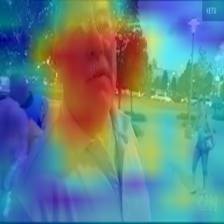}
\end{subfigure}

\begin{subfigure}{0.28\linewidth}
\includegraphics[width=\linewidth]{image/AJJFTGGI_000233_annotation.png}
\end{subfigure}
\hspace{0.5mm} 
\begin{subfigure}{0.28\linewidth}
\includegraphics[width=\linewidth]{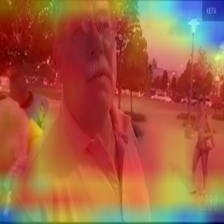}
\end{subfigure}
\hspace{0.5mm} 
\begin{subfigure}{0.28\linewidth}
\includegraphics[width=\linewidth]{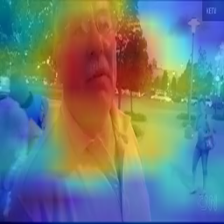}
\end{subfigure}

\caption{The first row is batch size 64, learning rate 5e-5, The second row is batch size 128, learning rate 3e-5,
The third row is batch size 128, learning rate 5e-5,}
\label{fig:semi result}
\vspace{-5mm} 
\end{figure}

In qualitative comparisons, we mainly visualize the localization results in figure \ref{fig:para result} and figure \ref{fig:semi result}.

We observe that when the scene is complex, the model is prone to overlook target objects (e.g., the first and second rows in Figure 3), while when the scene is simple (e.g., the first, second, and third rows in Figure 4), the model can locate target objects well but cover unrelated background details (e.g., ground and trees). It is hypothesized that because most of the occurring objects in the dataset are people, the model did not learn the vocal characteristics of the rest of the vocal objects (e.g., drums) very well. The semi-supervised model may have been disturbed by the data where some of the vocalized objects were not people, and the training effect was instead reduced. 

Similarly, the model tends to underestimate or overestimate the extent of the sound-emitting object, as evident in the first and second rows of Figure 2. This may be because positive and negative regions in different images cannot be easily distinguished using the same threshold parameter.

In contrast, the SSPL method can cover the regions of interest, and the use of PCM (possibly referring to a specific technique or method) further helps reduce the influence of background noise, resulting in more accurate localization.

\subsection{Quantitative Results and Analysis.}

\begin{figure}[h]
\centering
\begin{subfigure}{0.9\linewidth}
\caption*{learning rate = 3e-5}
\includegraphics[width=\linewidth]{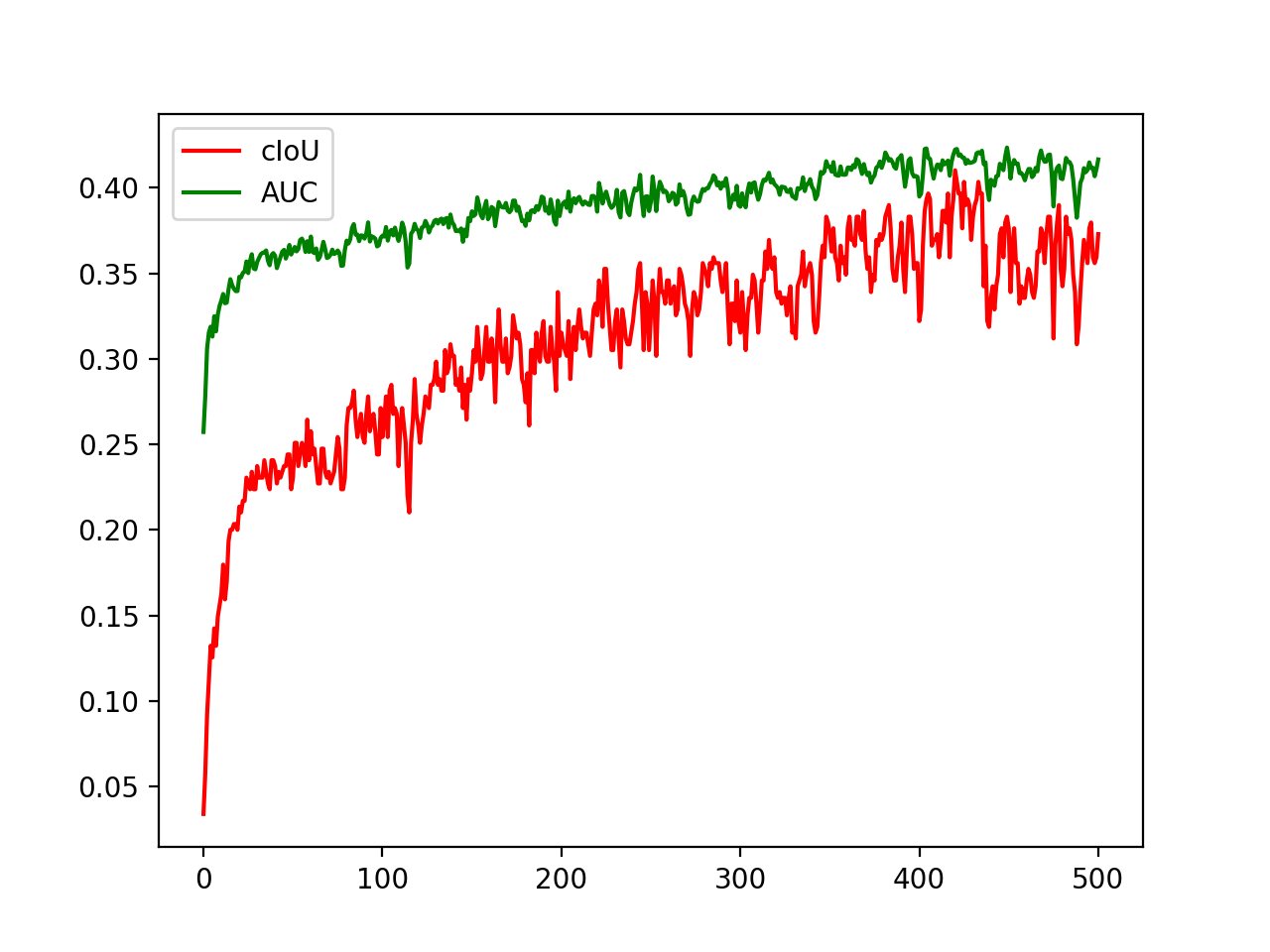}
\end{subfigure}
\hspace{0.5mm} 
\begin{subfigure}{0.9\linewidth}
\caption*{learning rate = 5e-5}
\includegraphics[width=\linewidth]{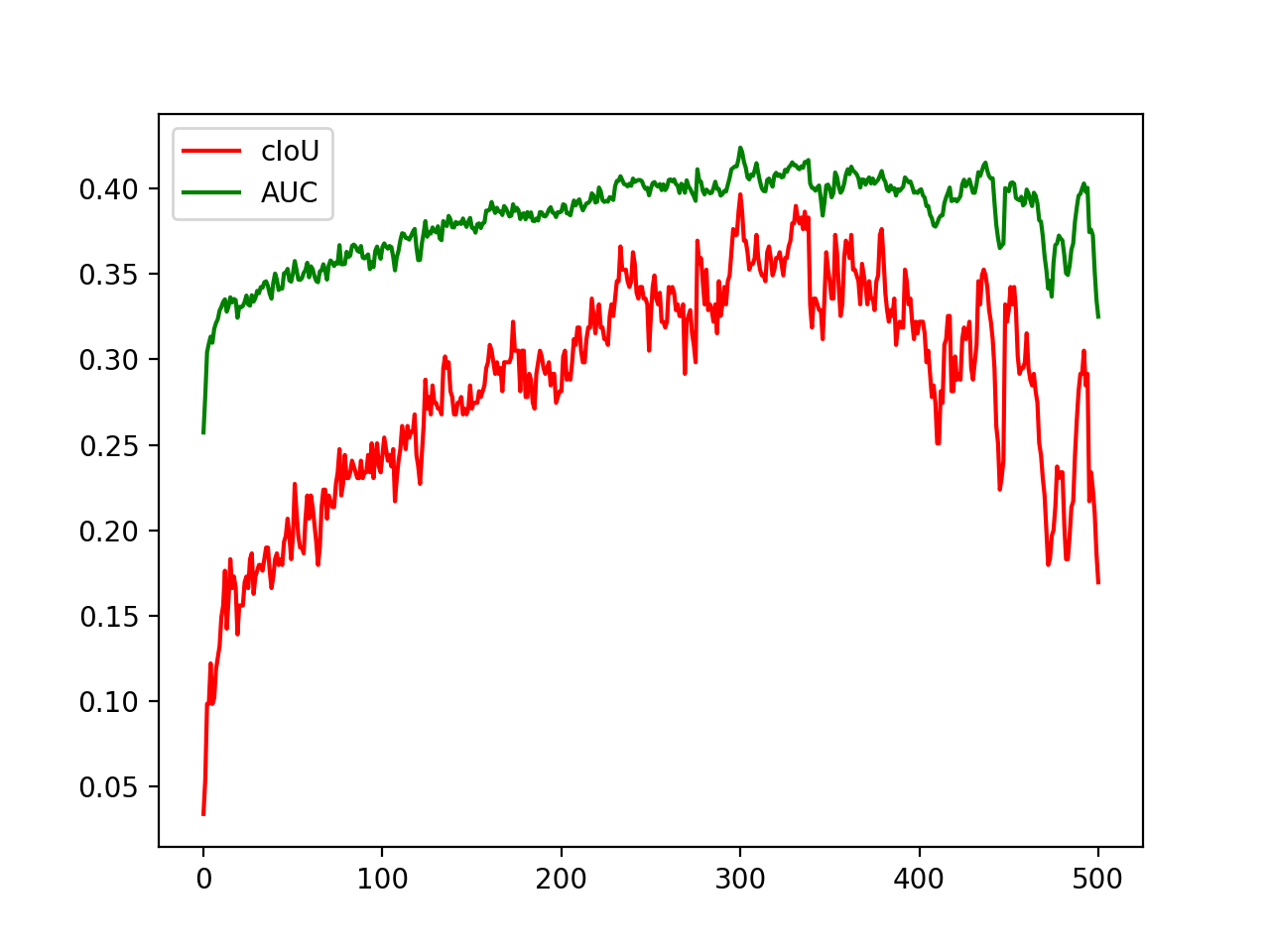}
\end{subfigure}

\caption{Different learning rate parameter results. The top learning rate is 3e-5, the bottom learning rate is 5e-5.}
\label{fig:ciou result}
\vspace{-5mm} 
\end{figure}

Table \ref{tab:para} shows results using different parameters, the best outcome is observed when batch size = 128. Larger batch size improves memory utilization as well as parallelization efficiency for large matrix multiplications, requires fewer iterations to run through an epoch (the full dataset), and is processed faster than a smaller batch size for the same amount of data. Within a certain range, generally speaking, the larger the batch size is, the more accurate the direction of descent it determines, and the less training oscillations it causes.
\begin{table}[h]
\begin{center}
\caption{Results of Parameter Adjustments} 
\label{tab:para}
\begin{tabular}{c c c c c}
  \hline
  Method & bz & lr & cIoU & AUC 
  \\
  \hline
  SSPL(Unsupervised) & 128 & 3e-5 &  41.02 & 42.23\\
  SSPL(Unsupervised) & 64 & 5e-5 & 42.03 & 43.38 \\
  SSPL(Unsupervised) & 128 & 5e-5 &  47.70 & 44.14\\
  \hline
\end{tabular}
\end{center}
\end{table}
As depicted in Figure \ref{fig:ciou result}, when examining the impact of varying learning rates on the training process, a noticeable trend emerges. Decreasing the learning rate appears to mitigate the fluctuations in accuracy throughout the training epochs. However, this seemingly beneficial effect comes at a cost: a significant reduction in the convergence speed of the model. In practical terms, this slowdown in convergence renders the training process prohibitively slow, thereby undermining the overall efficacy of the learning algorithm. Consequently, striking a balance between stability and speed becomes paramount in optimizing the learning rate for the given task or model architecture.

\begin{table}[h]
\begin{center}
\caption{Results of different kinds of supervision} \label{tab:semi}
\begin{tabular}{c c c}
  \hline
  Methods & cIoU & AUC
  \\
  \hline
  LSS(Unsupervised) &	23.85 &	36.56\\
  SemiPL(Semi-supervised) & 36.84 & 41.86\\
  SSPL(Unsupervised) & 47.70 & 44.14\\
  \hline
\end{tabular}
\end{center}
\end{table}

The performance of our SemiPL, while promising, is not quite on par with that of self-supervised models. One potential reason for this could be the relatively large bounding box range used during training. This extensive range may have interfered with the model's ability to learn precise localization, as it introduced too much noise and irrelevant information into the learning process. This hypothesis suggests that refining the bounding box or exploring new data annotation strategies may improve the performance of semi-supervised learning models. Future research in this area is expected to provide valuable insights and advances in the overall development of semi-supervised learning.

\section{Conclusion and Future Work}
In this study, we implemented the Self-Supervised Predictive Learning method (SSPL) and our novel Semi-supervised Predictive Learning approach, SemiPL, on the Chaotic World dataset. Our primary objective was to elevate the accuracy of visual-audio localization by explicitly mining positive correspondences. To achieve this, we utilized a tri-stream network coupled with a strategic training regimen to establish the relationship between audio signals and their corresponding video frames within the same video clip. Although SSPL proved effective in single-source audio localization tasks, it fell short in multi-source scenarios primarily due to the limited size of the dataset. Despite this limitation, the expanding availability of data and the escalating importance of label quality indicate that self-supervised learning is destined to become a prevailing trend in machine learning. A potential solution is to develop a fully supervised approach to multi-source localization tasks, which we leave for future work.

\section{Acknowledgement}
This work was supported by National Natural Science Foundation of China (No. 62203476), Natural Science Foundation of Shenzhen (No. JCYJ20230807120801002).

\bibliographystyle{IEEEbib}
\bibliography{icme2023template}

\end{document}